\documentclass[10pt,twocolumn,letterpaper]{article}

\usepackage[pagenumbers]{iccv} 

\definecolor{iccvblue}{rgb}{0.21,0.49,0.74}
\usepackage[pagebackref,breaklinks,colorlinks,allcolors=iccvblue]{hyperref}
\usepackage{times}
\usepackage{epsfig}
\usepackage{graphicx}
\usepackage{amsmath}
\usepackage{amssymb}
\usepackage{url}
\usepackage{algpseudocode}
\usepackage{algorithm}
\usepackage{multirow}
\usepackage{diagbox}
\usepackage{bm}
\usepackage{bbm}
\usepackage{mathrsfs}
\usepackage{enumitem}
\usepackage{caption}
\usepackage{colortbl}
\usepackage{booktabs}
\usepackage[numbers]{natbib}

\definecolor{lightgray}{gray}{0.92}   
\DeclareMathOperator*{\argmax}{arg\,max}
\DeclareMathOperator*{\argmin}{arg\,min}

\def\paperID{11169} 
\def\confName{ICCV}
\def\confYear{2025}

\title{PBCAT: Patch-Based Composite Adversarial Training against Physically Realizable Attacks on Object Detection}

\author{Xiao Li$^{1*}$  \ \ Yiming Zhu$^{1, 2*}$ \ \ Yifan Huang$^{1, 2}$\thanks{Equal contribution.} \ \ Wei Zhang$^{1}$ \ \ Yingzhe He$^{3}$ \ \ Jie Shi$^{3}$ \ \ Xiaolin Hu$^{1, 4}$\thanks{Corresponding author.}\\
$^{1}$
Department of Computer Science and Technology, BNRist, \\ 
IDG/McGovern Institute for Brain Research, THBI, Tsinghua University \\
$^{2}$University of Science and Technology Beijing \\
$^{3}$Huawei Technologies \quad $^{4}$Chinese Institute for Brain Research (CIBR)  \\
\tt\small \{lixiao20, zhang-w19\}@mails.tsinghua.edu.cn, \tt\small \{U202241171, U202240988\}@xs.ustb.edu.cn, \\ 
\tt\small \{heyingzhe, shi.jie1\}@huawei.com, \tt \small xlhu@mail.tsinghua.edu.cn
}

\begin{document}

\maketitle

\begin{abstract}
Object detection plays a crucial role in many security-sensitive applications. However, several recent studies have shown that object detectors can be easily fooled by physically realizable attacks, \eg, adversarial patches and recent adversarial textures, which pose realistic and urgent threats. Adversarial Training (AT) has been recognized as the most effective defense against adversarial attacks. 
While AT has been extensively studied in the $l_\infty$ attack settings on classification models, 
AT against physically realizable attacks on object detectors has received limited exploration. 
Early attempts are only performed to defend against adversarial patches, leaving AT against a wider range of physically realizable attacks under-explored.
In this work, we consider defending against various physically realizable attacks with a unified AT method. 
We propose PBCAT, a novel Patch-Based Composite Adversarial Training strategy. PBCAT optimizes the model by incorporating the combination of small-area gradient-guided adversarial patches and imperceptible global adversarial perturbations covering the entire image. With these designs, PBCAT has the potential to defend against not only adversarial patches but also unseen physically realizable attacks such as adversarial textures.
Extensive experiments in multiple settings demonstrated that PBCAT significantly improved robustness against various physically realizable attacks over state-of-the-art defense methods. Notably, it improved the detection accuracy by 29.7\% over previous defense methods under one recent adversarial texture attack\footnote{
Code is available at \url{https://github.com/LixiaoTHU/oddefense-PatchAT}}.
\end{abstract}

\section{Introduction}

\label{sec:intro}
Object detection, which requires simultaneously classifying and localizing all objects in an image, is a fundamental task in computer vision. Recent advancements in object detection methods \cite{odreview, frcnn, fcos, dndetr} have greatly benefited from the utilization of Deep Neural Networks (DNNs). However, DNNs are known to be susceptible to adversarial examples \cite{FirstExample} crafted by adding deliberately designed perturbations to the original examples. 

Adversarial examples exist not only in the digital world but also in the physical world \cite{AdvPatch, yolovpatch, tshirt, wu2020adv-cloak, AdvTexture, AdvCaT}. Several studies have demonstrated that object detectors can be easily fooled by \textit{physically realizable attacks}, \eg, adversarial patches \cite{AdvPatch, yolovpatch} and adversarial textures \cite{AdvTexture, AdvCaT} \footnote{Both adversarial patches and adversarial textures can be implemented in the physical world and are thereby called physically realizable attacks.}. Specifically, adversarial \textit{patch attacks} craft localized adversarial patterns within a fixed region (\eg, a square patch), while adversarial \textit{texture attacks} craft more pervasive adversarial perturbations that spread across the entire surface of the object, \eg, adversarial modifications to clothing textures that cover most of the surface of an object. Given the crucial role of object detection in numerous security-sensitive real-world applications, such as autonomous driving \cite{AutoDriving} and video surveillance \cite{video}, it is imperative to improve the adversarial robustness of object detectors against these physically realizable attacks, which pose realistic and severe threats.

Numerous studies \cite{locationPatchAT, locationiclr, metaAT, FNC, epgf, sac, ape, LGS, frame} have proposed various defense methods against physically realizable attacks. However, three critical limitations still hinder their effectiveness: 1) Most existing defenses are designed specifically for patch attacks, which represent only the simplest form of physically realizable attacks. Recent adversarial texture attacks \cite{AdvTexture, AdvCaT}, which generate adversarial clothing to deceive person detectors in the physical world, introduce a distinct threat model that differs from adversarial patches. To the best of our knowledge, defenses against such attacks remain largely unexplored. 2) Existing methods may be circumvented by specially designed techniques, \ie, adaptive attacks \cite{adaptive18, adaptive20}. Adversarial Training (AT) \cite{PGD} is widely recognized as an effective defense against adaptive attacks and has been extensively studied in the $l_\infty$ attack setting (referred to as \textit{$l_\infty$-AT}) \cite{PGD, trades, advod, zhang2019mtd, rock, cwat, zeroshot, adbm}. However, $l_\infty$ attacks involve global adversarial perturbations applied to all image pixels, which are impractical in the physical world. Thus, $l_\infty$-AT, which is designed for a fundamentally different threat model, may perform poorly against physically realizable attacks. 3) AT with adversarial patches (termed \textit{patch-based AT}) is a promising approach for defending against patch attacks. However, only a few early works \cite{locationiclr, locationPatchAT, metaAT} have explored its application in classification tasks. Extending these methods to object detection incurs prohibitive computational costs, which we discuss in \cref{sec:gradientpatch}.

To address these limitations, we consider defending against various physically realizable attacks with a unified AT method. Specifically, we introduce PBCAT, a novel Patch-Based Composite Adversarial Training strategy. First, we extend $l_\infty$-AT to patch-based AT, to improve robustness against adversarial patch attacks. 
Second, we propose a gradient-guided patch partition and selection method to efficiently find effective patch locations for AT.
Third, we incorporate global imperceptible adversarial perturbations used in $l_\infty$-bounded AT into patch-based AT, which primarily employs small-area, gradient-guided adversarial patches. By leveraging composite perturbations, PBCAT mitigates overfitting to patch attacks, enabling it to defend not only against adversarial patches but also against unseen physically realizable attacks, such as adversarial texture attacks with large-area perturbations. Finally, to enhance the practical utility of PBCAT, we draw inspiration from FreeAT \cite{freeAT} to enable PBCAT to train a robust detector at a computational cost comparable to standard training.

We trained object detectors with PBCAT on the MS-COCO \cite{coco} dataset. Evaluations were performed on several datasets, including the MS-COCO dataset for the general object detection task as well as the Inria \cite{inria} dataset along with a synthetic dataset \cite{AdvTexture} for the downstream security-critical person detection task. We demonstrated that PBCAT significantly improved robustness against various physically realizable attacks over state-of-the-art (SOTA) defense methods in strong adaptive settings \cite{adaptive18, adaptive20}. On the person detection task, PBCAT secured a Faster R-CNN \cite{frcnn} with 60.2\% and 56.4\% Average Precision (AP) against two recent adversarial texture attacks, AdvTexture \cite{AdvTexture} and AdvCaT \cite{AdvCaT}, respectively. Notably, PBCAT achieved a 29.7\% AP improvement over the SOTA defense methods against AdvTexture.

Our main contributions can be summarized as follows:
\begin{itemize}
    \item We propose PBCAT, a novel adversarial training method that unifies the defense against various physically realizable attacks within a single model;
    \item PBCAT closes the gap between adversarial patches and adversarial textures by employing composite perturbations along with patch partition and gradient-guided selection techniques;    
    \item Extensive experiments demonstrate that PBCAT achieves superior adversarial robustness against diverse physically realizable attacks in strong adaptive settings.
\end{itemize}

\section{Preliminary and Related Work}
\label{sec:related}

\subsection{Adversarial Robustness on Classification}
Adversarial examples, first discovered on image classification \cite{FirstExample}, are input images with deliberately designed perturbations that can fool DNN-based image classifiers while still being easily recognized by humans. Given an image-label pair $(\mathbf{x}, y)$ and a classifier $f_\theta(\cdot)$, adversarial perturbation $\mathbf{\delta}$ can be found by maximizing the output loss: $\mathbf{\delta} = \argmax_{\mathcal{B}(\mathbf{\delta}) \leq \epsilon} \mathcal{L}(f_\theta(\mathbf{x} + \mathbf{\delta}), y)$, where $\mathcal{L}$ denotes a cross-entropy loss, and the attack intensity $\epsilon$ bounds the attack budget $B$. Several approximate methods \cite{PGD, FGSM, cw} have been proposed to solve the intractable maximizing problem. PGD \cite{PGD} is one of the most popular methods, which optimizes perturbations through multiple iterations with small step sizes. AT and its variants are generally considered the most effective defense methods against adversarial examples, which improve adversarial robustness by incorporating adversarial examples into training:
\begin{equation}
\begin{aligned}
  \theta = \argmin_{\theta}\mathbb{E}_{\mathbf{x}}\{\max_{\mathcal{B}(\mathbf{\delta}) \leq \epsilon} \mathcal{L}(f_\theta(\mathbf{x} + \mathbf{\delta}), y)\}.
\end{aligned}
\label{eq:at}
\end{equation}
However, most works investigate AT on classification models in the $l_\infty$-bounded settings \cite{PGD, trades, zeroshot}, \ie, $\mathcal{B}(\cdot):=\|\cdot\|_\infty$, which involve adding a global adversarial perturbation to the images and are generally considered as physically infeasible attacks. 

\subsection{Adversarial Attacks on Object Detection}
Object detection requires simultaneously classifying and localizing all objects in an image. Modern object detectors \cite{odreview, frcnn, fcos} have significantly improved with the utilization of DNNs. Object detectors are also vulnerable to adversarial examples, and several works \cite{failat, fabattack, cwat} have investigated how to attack object detectors from various aspects. Unlike classifiers where $l_\infty$ attacks and defenses are often investigated, more urgent physically realizable attacks are widely studied for this task, considering that object detection has been widely used in many security-critical applications. Particularly, for the person detection task, several physically realizable adversarial attacks have been proposed. \citet{AdvPatch} first propose to generate physically realizable adversarial patches to fool person detectors, which we denote AdvPatch. AdvTexture attack \cite{AdvTexture} further extends AdvPatch to tileable adversarial textures, offering attack effects from various viewing angles. It proposes a scalable generative method to craft adversarial texture with repetitive structures. AdvCaT attack \cite{AdvCaT} optimizes adversarial textures into typical camouflage patterns to resemble cloth patterns in the physical world. Attackers can print the clothing texture created by AdvCaT and AdvTexture on a piece of cloth and tailor it into an outfit. Wearing such an outfit can hide the person from SOTA detectors, posing realistic and urgent security threats. 

\subsection{Adversarial Defenses against Patch Attacks}
\label{sec:patchat}
To defend against adversarial examples of object detectors, especially patch-based attacks, several types of methods have been proposed. Input preprocessing-based methods, such as LGS \cite{LGS}, SAC \cite{sac}, EPGF \cite{epgf}, and Jedi \cite{jedi}, mask out or suppress the potential adversarial patch areas before sending them to the model. 
Outlier feature filter-based methods, such as FNC \cite{FNC} and APE \cite{ape}, incorporate filters into the model to smooth abnormal inner features caused by adversarial patches. Defensive frame methods, such as UDF \cite{frame}, train an adversarial defense frame surrounding images to improve robustness. However, similar to the experiences on $l_\infty$ defense \cite{adaptive18, adaptive20}, These non-AT methods can be vulnerable to strong adaptive attacks, as detailed in our experimental results in \cref{sec:results}. 
Besides these empirical methods, some studies \cite{certified, certifiedod} investigate to improve certified robustness, but till now the certified methods only work under quite tiny perturbations and need time-consuming inference. In this work, we mainly compare PBCAT with empirical methods in practical scenarios.

Only a few early works \cite{locationiclr,locationPatchAT,metaAT} have explored patch-based AT for defending against patch attacks. Both \citet{locationPatchAT} and \citet{locationiclr} investigate methods for identifying optimal patch locations for patch-based AT in classification models, while \citet{metaAT} propose an enhanced patch-based AT approach based on meta-learning. Although these methods demonstrate promising results against patch attacks, they were not originally designed for object detection, and adapting them to this task presents significant challenges. For instance, \citet{locationPatchAT} and \citet{locationiclr} primarily focus on selecting a single optimal patch position for classification, which typically involves a single object. In contrast, object detection requires handling multiple bounding boxes, necessitating the identification of multiple effective patch locations. As a result, directly applying these methods to object detection leads to a substantial increase in computational complexity. Moreover, relying solely on adversarial patches for training does not generalize well to a broader range of physically realizable attacks, as discussed later.

\section{PBCAT}
\label{sec:method}
PBCAT aims to defend against various physically realizable attacks by incorporating a combination of a small area of sub-patches and a large area of imperceptible adversarial perturbations. We discuss how to build patch-based AT methods from $l_\infty$-AT in \cref{sec:patchat_method}. We then describe the gradient-guided patch partition and selection method in \cref{sec:gradientpatch}. In \cref{sec:composite}, we introduce how to extend patch-based AT to defend against large-area perturbations. Finally, we show how to accelerate PBCAT in \cref{sec:freeat}.

\begin{figure*}[!t]
\centering
\includegraphics[width=0.75\linewidth]{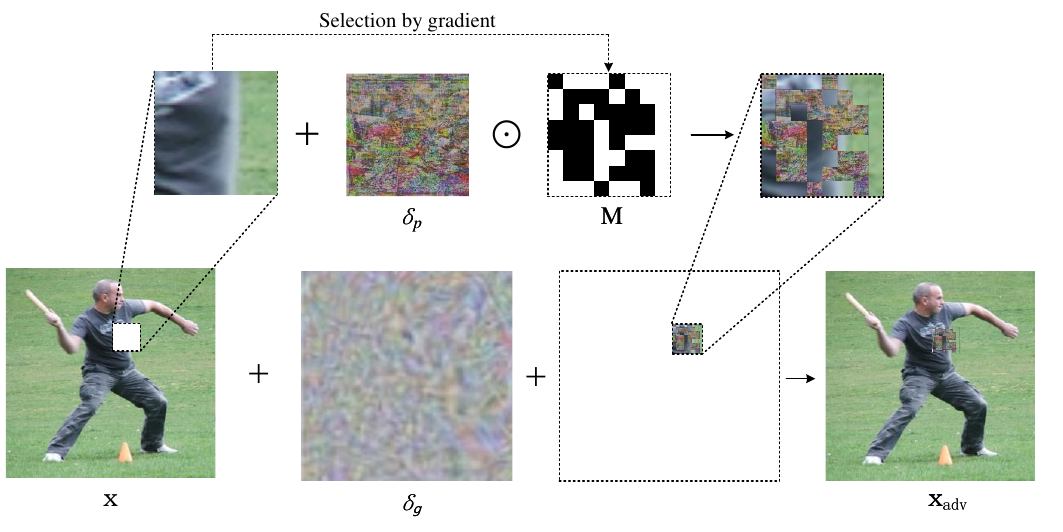}
\vspace{-4mm}
\caption{The illustration on how to generate training images for PBCAT. A patch location is randomly selected from each bounding box area first. The selected patch is then partitioned into multiple sub-patches. The average gradient norm is computed for each partitioned sub-patch, and the top half with the highest values are selected to construct a binary mask $\mathbf{M}$. This mask is then applied element-wise to the patch perturbation $\delta_p$, forming a new masked patch. The global small perturbation $\delta_g$ and these sub-patches are added to obtain the final adversarial example $\mathbf{x}_\mathrm{adv}$. Since $\delta_g$ is constrained by $\|\delta_g\|_\infty \leq 4/255$, we scale $\delta_g$ in the visualization for better clarity.}

\label{fig:pbcat}
\end{figure*}

\subsection{From $l_\infty$-AT to Patch-Based AT}
\label{sec:patchat_method}

Given an image $\mathbf{x} \in [0, 1]^{3\times H\times W}$ and its corresponding bounding box labels $y$, where $H \times W$ denotes the input resolution, $l_\infty$ attacks typically constrain the perturbation budget $\mathcal{B}$ with a bound $\epsilon$: $\|\cdot\|_\infty \leq \epsilon$. In practice, $\epsilon$ is often set to a small value, \eg, $4/255$ \cite{advod, transfer}. However, $l_\infty$-bounded perturbations apply \textit{global} modifications to all pixels in $\mathbf{x}$, which is generally considered physically unrealistic. In contrast, physically realizable attacks restrict perturbations to \textit{local} regions, typically targeting the foreground of an object \cite{AdvPatch, AdvTexture, AdvCaT}, where pixel modifications can have a higher intensity. Unlike classification tasks, object detection involves multiple foreground objects within a single input image. In this work, we adopt a threat model where each bounding box may contain an adversarial patch. Since $l_\infty$ attacks and physically realizable attacks assume different perturbation constraints, and AT often exhibits suboptimal performance against unseen threats \cite{unseen1, unseen2}, we opt for patch-based AT as a more effective defense against physically realizable attacks.

On the other hand, according to \cref{eq:at}, patch-based AT follows a similar formulation to $l_\infty$-AT, with the key distinction being the introduction of a mask to restrict perturbations to specific regions. Specifically, patch-based AT for object detectors can be formulated as follows:

\begin{equation}
\small
\label{eq:bound}
\theta = \argmin_{\theta}\mathbb{E}_{\mathbf{x}}\{\max_{\|\delta_p \odot \mathbf{M}\|_\infty \leq \beta} \mathcal{L}_d(f_\theta(\mathbf{x} + \delta_p \odot \mathbf{M}), y)\},
\end{equation}
where $\delta_p$ denotes the patch perturbation, while $\mathbf{M}$ denotes a binary mask for restricting $\delta_p$ to a local area. $\beta$ denotes a large perturbation intensity (\eg, $1$), and $\mathcal{L}_d$ denotes the loss of an object detector. The inner maximizing problem of patch-based AT can take inspiration from $l_\infty$-AT, as detailed in \cref{sec:freeat}. We discuss how to obtain $\mathbf{M}$ next.

\subsection{Instance-Level Patch Placement} 
\label{sec:bboxpatch}


For object detection tasks, the adversarial patch is typically created and placed to a fixed location relative to the bounding box. But adopting this approach in AT may cause information leakage for bounding box prediction, as models might utilize the adversarial patch to perform bounding box regression. PBCAT randomly samples adversarial patch locations within a bounding box to mitigate this issue. Denoting the size of the bounding box as $(\mathit{w}_{\mathrm{bbox}}, \mathit{h}_{\mathrm{bbox}})$, and the center of the bounding box as $(\mathit{x}_{\mathrm{bbox}}, \mathit{y}_{\mathrm{bbox}})$, The center $(x_\mathrm{s}, y_\mathrm{s})$ of the patch is obtained by randomly sampling patch locations within the bounding box via a Gaussian distribution with $(\mathit{x}_{\mathrm{bbox}}, \mathit{y}_{\mathrm{bbox}})$ as the mean. The width and height of the sampled patch are set to be $s=\lambda\cdot\sqrt{w_{\mathrm{bbox}}^2+h_{\mathrm{bbox}}^2}$, where $\lambda$ is a hyper-parameter. 

\begin{figure*}[!t]
\centering
\includegraphics[width=0.8\linewidth]{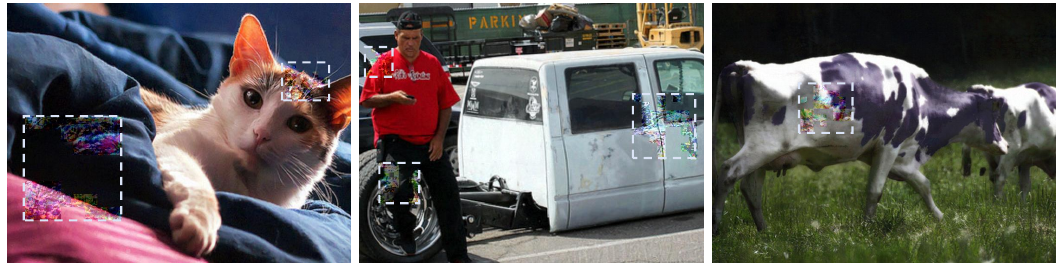}
\vspace{-2mm}
\caption{Visualization of training examples with adversarial perturbations used in PBCAT. Perceptible gradient-guided adversarial patches and imperceptible global adversarial perturbations are added to each image. The patch regions are annotated with white dashed boxes.}
\label{fig:train}
\end{figure*}

\subsection{Gradient-Guided Patch Partition and Selection} 
\label{sec:gradientpatch}

As highlighted in early patch-based AT attempts \cite{locationPatchAT, locationiclr}, patch location plays a crucial role in enhancing the effectiveness of patch-based AT. However, these methods often rely on time-consuming trial-and-error processes, such as multiple forward inferences to determine optimal patch locations \cite{locationPatchAT} within a specific optimization step. This issue is exacerbated in object detection, where an image typically contains multiple objects, leading to a computational overhead that scales with the number of candidate locations. Thus, extending these methods to object detection incurs prohibitive theoretical computational costs, which we further analyze in \cref{appen:computation}. 

In contrast, PBCAT efficiently determines patch locations using gradient-based selection, requiring only a single forward/backward pass for each optimization step. 
For simplicity, here we consider the case of a single bounding box, noting that the multi-object scenario follows a similar process. Given a randomly placed patch (as described in \cref{sec:bboxpatch}), we partition it into $n \times n$ sub-patches, resulting in $N = n^2$ partitioned areas, where $N$ is a hyper-parameter. After a single backward pass to obtain the whole gradient information for this patch, we compute the average gradient norm for each sub-patch and select the top 50\% with the highest values, as the areas with large gradient norms generally are the vulnerable areas that have a significant impact on the output loss. For implementation, we define a binary mask $\mathbf{M}$ of the same size as the original patch. This mask, applied via element-wise multiplication, determines which regions of the patch are retained. An illustration of $\mathbf{M}$ is shown at the top of \cref{fig:pbcat}. Our method effectively identifies vulnerable regions for adversarial training using only a single forward and backward pass, incurring negligible additional computational cost compared to previous methods.

\subsection{Local Patches and Global Noises} 
\label{sec:composite}

Recent adversarial texture attacks \cite{AdvTexture, AdvCaT} adopt a significantly larger area attack than patch-based attack \cite{AdvPatch}. Training with small adversarial patches only makes it challenging to defend against these large-area physically realizable attacks (see \cref{sec:ablation}, where training with patches alone exhibited poor robustness against texture attacks). To defend against these attacks, a direct and ideal method would be to perform AT with large-area unrestricted adversarial noises. However, we find that simply increasing the patch size (perturbation area) can induce training collapse of patch-based AT, \ie, the large-area unrestricted perturbation incurred slow training convergence and poor robustness (see \cref{tab:patchsize}, where doubling the patch size significantly reduced robustness against various attacks). We guess that the collapse might be caused by the model’s limited capacity and significant corruption of object information (a large-area unrestricted patch perturbation can corrupt the information of the entire object).

Instead, we propose to incorporate global imperceptible adversarial perturbations generally used in $l_\infty$-AT into the patch-based AT. The insight behind this approach is as follows: 1) By incorporating large-area adversarial noises while constraining the attack intensity ($l_\infty$-bound), sufficient object information can be kept to avoid training collapse. 2) On the other hand, since the perturbation areas of unseen physically realizable attacks may be located at arbitrary locations of an image (\eg, using a printable patch), the $l_\infty$-bounded global noise ensures the entire image to be covered by adversarial perturbations during training. 3) Training with $l_\infty$-bounded noise has been shown to be helpful against several different adversarial threats \cite{revisiting, pinpp}, and our subsequent results further confirms that $l_\infty$-AT can enhance robustness against patch attacks. The final perturbation used in PBCAT is:
\begin{equation}
\label{eq:A}
\delta = \mathrm{Apply}(\delta_p \odot \mathbf{M}, \mathbf{x}) + \delta_{g},
\end{equation}
where $\delta_{p}$ denotes the patch perturbation, $\delta_{g}$ denotes the global noises, \ie, $\|\delta_{g}\|_\infty \leq \epsilon$, and $\mathrm{Apply}$ denotes the operation of pasting the patch back onto the image $\mathbf{x}$.

\subsection{Accelerating AT} 
\label{sec:freeat}

Early patch-based AT works \cite{locationiclr, locationPatchAT, metaAT} employ full PGD attacks to solve the inner maximization problem of AT and train object detectors from scratch, making the process highly time-consuming. In PBCAT, inspired by recent SOTA $l_\infty$-AT practices \cite{advod} for object detection, we adopt FreeAT \cite{freeAT} as the default setting for patch-based AT and utilize an adversarially pre-trained backbone network \cite{advod}. FreeAT recycles gradient perturbations to mitigate the additional training cost introduced by inner maximization while maintaining comparable adversarial robustness. We initialize both the patch perturbation $\delta_p$ and the global perturbation $\delta_g$ to zero. At each iteration, we compute the gradient, take its sign, and update $\delta_p$ and $\delta_g$ using different step sizes. The pseudo-code of ``Free'' PBCAT for object detection is provided in \cref{appen:alg}. Notably, the adversarial gradient $\mathbf{g}_{\mathrm{adv}}$ and model parameter gradient $\mathbf{g}_{\theta}$ are computed simultaneously in a single backward pass. By leveraging ``Free'' PBCAT, the training cost of adversarial training becomes comparable to standard training. The actual training time is detailed in \cref{appen:setting}.

\begin{table*}[!t]
  \centering
  \small
  \setlength{\tabcolsep}{4pt}
  {

    \begin{tabular}{c|cc|ccc}
    \toprule
     Method & Clean (Inria) & AdvPatch & Clean (Synthetic) & AdvTexture & AdvCaT \\
     \midrule
        Vanilla & $96.2$ & $37.3$ & $86.4$ & $0.2$ & $0.3$\\
	 \hline
        LGS \cite{LGS} &$95.9$& $24.1$ & $86.4$ &$3.9$ & $4.3$  \\
        SAC \cite{sac} &$96.2$& $57.1$ & $85.4$ & $0.3$ & $0.6$  \\
        EPGF \cite{epgf} &$95.1$& $43.2$ & $86.7$ & $2.9$ & $0.4$  \\
        Jedi \cite{jedi} & $92.3$ & $64.4$ & $88.1$ & $2.3$ & $0.7$ \\
        FNC \cite{FNC} &$96.8$& $53.0$ & $81.9$ & $6.0$ & $5.8$  \\
        APE \cite{ape} &$96.2$& $47.9$ & $81.9$ & $0.8$ & $0.3$  \\
        UDF \cite{frame} &$69.1$& $19.3$ & $84.9$ & $2.2$ & $5.8$  \\
        PatchZero \cite{PatchZero} &$96.2$& $38.5$ & $79.4$ & $0.0$ & $0.2$ \\
        NAPGuard \cite{NAPGuard} &$96.1$& $47.0$ & $81.1$ & $2.2$ & $0.4$  \\
        AD-YOLO \cite{adyolo} &$90.0$& $39.0$ & $90.2$ & $0.5$ & $0.2$  \\
	\hline
       $l_\infty$-AT (MTD) \cite{zhang2019mtd} & $92.3$ & $54.6$ & $89.4$ & $1.4$ & $34.5$ \\
        $l_\infty$-AT (AdvOD) \cite{advod} &$95.9$& $56.1$ & $92.5$ & $30.5$ & $39.6$  \\
        \rowcolor{lightgray} PBCAT (Ours) &$95.4$& $\mathbf{77.6}$ & $92.5$ & $\mathbf{60.2}$ & $\mathbf{56.4}$ \\
     \bottomrule
    \end{tabular}
     \vspace{-2mm}
    }
    \caption{The detection accuracies (AP$_{50}$) of models with different defense methods under adaptive attacks. Clean (Inria) and AdvPatch \cite{AdvPatch} were evaluated on the Inria dataset. Clean (Synthetic), AdvTexture \cite{AdvTexture}, and AdvCaT \cite{AdvCaT} were evaluated on the synthetic dataset. }   
  \label{tab:downstream}
\end{table*}

\begin{figure*}[!t]
\centering
\includegraphics[width=0.79\linewidth]{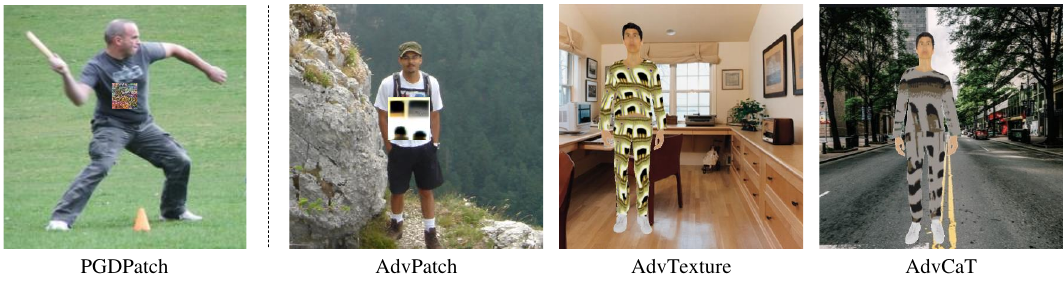}
\vspace{-2mm}
\caption{Visualization of examples of four attacks used in our evaluation. PGDPatch was optimized on MS-COCO. AdvPatch was optimized on the Inria dataset. AdvTexture and AdvCaT were optimized on the synthetic dataset used in \cite{AdvCaT}. PGDPatch is used to evaluate the general object detection task while other attacks were used to evaluate the downstream person detection.}
\label{fig:attack}
\end{figure*}

\section{Experiments}
\label{sec:exper}

\subsection{Experimental Settings}
\label{sec:setting}

Unless otherwise specified, our primary experiments were conducted on the widely used two-stage detector Faster R-CNN \cite{frcnn} with a ResNet-50 \cite{resnet} backbone. Additionally, we evaluated PBCAT on the one-stage detector FCOS \cite{fcos} and the recent DN-DETR \cite{dndetr}, with results presented in \cref{sec:more_detectors}. Notably, Faster R-CNN, FCOS, and DN-DETR have distinct architectures. We did not include YOLO \cite{yolov3, yolov8} series detectors in this study, as discussed in \cref{appen:setting}. PBCAT was trained on a diverse set of general objects. However, since most physically realizable attacks \cite{AdvPatch, yolovpatch, AdvTexture, AdvCaT} target security-critical applications, such as person detection, we mainly evaluated PBCAT-trained detectors on the person detection task.

\noindent\textbf{Datasets and metrics.} Three datasets are used in this work: 1) The MS-COCO \cite{coco} dataset for training the general object detectors. We used the 2017 version, containing 118,287 images of 80 object categories for training and 5,000 images for evaluation. 2). The Inria Person \cite{inria} dataset for the person detection task. Note that this dataset is only used for attack evaluation. 3) The synthetic dataset used in \citet{AdvTexture, AdvCaT} for evaluation. This dataset contains 506 background images of different scenes, with 376 images used for adversarial texture optimization and 130 images for evaluation. We rendered a 3D person wearing an adversarial outfit optimized by AdvTexture and AdvCaT on the provided background images using a differential renderer \cite{renderer}. We used AP$_{50}$, AP with an IoU threshold of 0.5, as the primary metric for evaluating the detection accuracy under different attacks, considering that it is a widely-used and practical metric for object detection \cite{advod, yolov3}.

\noindent\textbf{Training recipe of PBCAT.} We trained the object detector with PBCAT on MS-COCO. Unless otherwise specified, the patches in each image were generated with a patch perturbation step size $\alpha = 8/255$, patch perturbation intensity $\beta = 64/255$, scale factor $\lambda = \sqrt{2}/5$, and the amount of sub-patches $N = 64$. The replay parameter for FreeAT was set to be $r = 8$. Specifically, each bounding box had a 50\% chance of being attached to an adversarial patch to increase the detection accuracy for objects without adversarial patches (clean objects). The global perturbations were generated with a perturbation intensity $\epsilon = 4/255$. Additional training settings basically followed the recipe proposed by \citet{advod} (see \cref{appen:setting}), which resulted in the recent SOTA robustness against the $l_\infty$ attack. Some training examples used in PBCAT are shown in \cref{fig:train}.

\noindent\textbf{Attack evaluation setups.}
For the general detection task on MS-COCO, our attack evaluation used the masked PGD attack to create the adversarial square patch for each bounding box, termed PGDPatch attack. Here the iteration step was set to 200, the step size was set to 2, and the hyper-parameter for the patch size was set to $\lambda = 1/5\sqrt{2}$, resulting in a patch area of 1\% to 5\% relative to the area of the bounding box. Note that the physical implementation of the patches created by PGDPatch was relatively difficult because the tricks for physical implementation like the TV loss \cite{tvloss} were not used. Instead, we cared more about the security-critical person detection task, where three actual physically realizable attacks were evaluated: AdvPatch \cite{AdvPatch}, AdvTexture \cite{AdvTexture}, and AdvCaT \cite{AdvCaT}. The evaluation settings for these three attacks strictly followed their original configurations in the digital world: The detector trained on MS-COCO was evaluated on the downstream person detection task directly. All of the three attacks also have their implementations and evaluations in the physical world. However, validating the effectiveness of defense methods against these physically realizable attacks in the digital world is sufficient, because the attack success rates of physically realizable attacks in the digital world is typically higher than those in the real physical world \cite{AdvPatch, AdvTexture}. In real physical world, physical implementation errors and differences in physical conditions (\eg, illumination) decrease the attack success rates. Thus, if a defense method performs well in defending against these physically realizable attacks in the digital world, it can exhibit stronger robustness in the real physical world. Similar evaluation paradigms have also been adopted by many defense methods \cite{locationiclr, epgf, sac}.

\subsection{Robustness against Adversarial Attacks} 
\label{sec:results}

We first compared PBCAT with the recent SOTA $l_\infty$-AT method for object detection, denoted as \textit{AdvOD} \cite{advod}, whose training strategy was also adopted in our work, on the general object detection task. The results shown in \cref{appen:coco} indicate that PBCAT achieved an AP$_{50}$ of $37.8\%$ averaged across all object categories and an AP$_{50}$ of $34.5\%$ for the person category under the PGDPatch attack, outperforming \cite{advod} by $6.1\%$ and $4.4\%$, respectively. These results demonstrate the potential of PBCAT in defending against physically realizable attacks in general object detection. We then turn to the actual physically realistic attacks on the security-critical person detection task.

In the person detection task, we compared PBCAT with various defense approaches against patch-based attacks, such as input preprocessing-based methods and outlier feature filter-based methods (see \cref{sec:patchat}). These defense methods were applied to the standardly trained detector, denoted as \textit{Vanilla}. We also compared PBCAT with more $l_\infty$-AT methods, including AdvOD \cite{advod} and MTD \cite{zhang2019mtd}. We give the implementation details of these baseline methods in \cref{appen:details}. Please note that early patch-based AT works \cite{locationiclr, locationPatchAT, metaAT} were not evaluated as these methods were proposed originally for the classification task and it is challenging to adapt to object detection (see \cref{sec:gradientpatch}). All of these defense methods were evaluated in the white-box adaptive setting to show the worst-case robustness. Note that PBCAT is an AT method that has full gradient available, so we employed the direct adaptive attack rather than specific techniques like BPDA \cite{adaptive18}.

The results of different defense methods are shown in \cref{tab:downstream}. These non-AT defense methods are all vulnerable to the adaptive patch attack. For the AdvPatch attack, the best non-AT defense method was SAC \cite{sac}, achieving 57.1\% AP$_{50}$. Moreover, against the stronger adaptive adversarial texture attacks, all these non-AT defense methods were broken. Interestingly, we observed that $l_\infty$-AT outperformed all non-AT defense methods, although the latter is explicitly designed for physically realizable attacks. Note that \citet{advod} did not evaluate their method against physically realizable attacks in their original work, as their focus was on defending against $l_\infty$-bounded perturbations. Nevertheless, our results show that $l_\infty$-AT  also enhanced robustness against physically realizable attacks. 
Compared with $l_\infty$-AT, PBCAT further improved the robustness. In particular, against AdvTexture, PBCAT improved the detection accuracy by 29.7\%.  \cref{appen:visualization} visualizes some detection results of the model trained with PBCAT under physically realizable attacks.

\begin{table}[!t]
  \centering
  \small
  \setlength{\tabcolsep}{4pt}
  {
    \begin{tabular}{ccc|ccc}
    \toprule
    Patch & Global & Gradient & AdvPatch & AdvTexture & AdvCaT \\
     \hline
     \checkmark & & & $35.4$ & $1.6$ & $0.8$ \\
    \checkmark& \checkmark & & $72.8$ & $24.9$ & $19.5$ \\
      \checkmark & \checkmark & \checkmark & $\mathbf{77.6}$ & $\mathbf{63.3}$ & $\mathbf{56.4}$ \\
      
     \bottomrule
    \end{tabular} 
    \vspace{-2mm} 
    }
    \caption{The detection accuracies (AP$_{50}$) of models trained with different ablation settings. AdvPatch \cite{AdvPatch} was evaluated on Inria; AdvTexture \cite{AdvTexture} and AdvCaT \cite{AdvCaT} were evaluated on the synthetic dataset used in \citet{AdvCaT}. 
    } 
  \label{tab:ablation}
\end{table}

\begin{table}[!t]
  \centering
  \small
  \setlength{\tabcolsep}{4pt}
  {
    \begin{tabular}{c|ccc}
    \toprule
    Method & AdvPatch & AdvTexture & AdvCaT \\
     \hline
     Random & $59.0$ & $14.2$ & $47.0$ \\
       Gradient & $\mathbf{77.6}$ & $\mathbf{63.3}$ & $\mathbf{56.4}$ \\
      
     \bottomrule
    \end{tabular}  
    \vspace{-2mm}
    }
    \caption{The selection strategies for sub-patches. ``Random'' represents selecting half of the sub-patches randomly. ``Gradient'' selects the sub-patches with the top $50\%$ highest gradient values.}
  \label{tab:selection}
\end{table}

Despite the distinct features between training examples (\cref{fig:train}) and the evaluated attack examples (\cref{fig:attack}), PBCAT model had strong robustness against various attacks, showing its good transferability across different attacks.
To further validate this, we evaluated two transfer-based patch attacks \cite{T-SEA, Nat-Patch} on our model, and the results detailed in \cref{appen:transfer} show that these transfer attacks almost lose the ability to fool the detectors trained with PBCAT.

\subsection{Ablation Study} 
\label{sec:ablation}

\subsubsection{Effectiveness of Each Design in PBCAT}

We first conducted an ablation study to show the effectiveness of each design in PBCAT: using the small area patch perturbations, using the global imperceptible noise perturbations, and using the gradient-guided patch partition and selection technique, denoted as ``Patch", ``Global", and ``Gradient", respectively. Using only ``patch'' indicates that its shape is a square.
The results are shown in \cref{tab:ablation}. We can see that all of these designs contribute to enhancing robustness against physically realizable attacks. By incorporating the global perturbations in conjunction with adversarial patches, the robust accuracy against the AdvPatch \cite{AdvPatch} attack increased to 72.8\%. Introducing the gradient-guided patch partitioning and selection technique further improved robustness across all attacks.


Additionally, we examined different sub-patch selection strategies, with results shown in \cref{tab:selection}. ``Random'' refers to selecting half of the sub-patches randomly, and ``Gradient'' selects the top $50\%$ with the highest gradient values. Notably, selecting sub-patches randomly after partitioning resulted in sub-optimal robustness, further highlighting the importance of gradient guidance in our selection process.

\subsubsection{Influence of Hyper-Parameters}

\noindent\textbf{The number of sub-patches.} Since the sampled large patch is partitioned into sub-patches, an exploration is required to identify the optimal number of partitions. We conducted three experiments by varying the number of sub-patches $N$: 16 ($4 \times 4$), 64 ($8 \times 8$), and pixel-level partitioning (each pixel as a sub-patch). Other settings followed those described in \cref{sec:setting}. \cref{tab:grad_downstream} demonstrates that the number of sub-patches requires a balance.

\noindent\textbf{The size of the sampled patch.} $\lambda$ controlled the size of the sampled patch, representing the ratio of the edge length of a square patch to the diagonal length of the bounding box. We consider three models with different $\lambda$ values to examine their effect. The results shown in \cref{tab:patchsize} indicate that simply enlarging the size of the sampled patch was ineffective when defending against large-area texture attacks. Moreover, too large sampled patches had negative effects even when defending against the patch-based attack.

\noindent\textbf{The selection of top-\textit{k} values.}
In \cref{sec:gradientpatch} we select the top 50\% of sub-patches with the highest norm values. Here we conducted experiments with varying top-\textit{k} values. The results in \cref{appen:topk} show that selecting the top 50\% strikes a favorable balance, yielding robust performance across different scenarios.

\begin{table}[!t]
    \centering
  \small
  \setlength{\tabcolsep}{4pt}
  {

    \begin{tabular}{c|ccc}
    \toprule
     Sub-patches & AdvPatch & AdvTexture & AdvCaT \\
     \hline
     16  & $\mathbf{78.3}$ & $50.8$ & $46.2$  \\
     \rowcolor{lightgray} 64  & $77.6$ & $\mathbf{60.2}$ & $56.4$  \\
     Pixel-level  & $67.4$ & $20.4$ & $\mathbf{59.4}$  \\
     \bottomrule
    \end{tabular}   
    \vspace{-2mm}
    }
    \caption{AP$_{50}$ of models trained with different numbers of sub-patches. Pixel-level means each pixel is treated as a sub-patch.}
    \label{tab:grad_downstream}%
\end{table}

\begin{table}[!t]
    \centering
  \small
  \setlength{\tabcolsep}{4pt}
  {

  \begin{tabular}{c|ccc}
    \toprule
     Scale factor $\lambda$  & AdvPatch & AdvTexture & AdvCaT \\
     \hline
     \rowcolor{lightgray} $2\sqrt{2}/10$ & $78.3$ & $\mathbf{50.8}$ & $\mathbf{46.2}$  \\
     $3\sqrt{2}/10$ & $\mathbf{80.4}$ & $49.1$ & $38.5$  \\
     $4\sqrt{2}/10$ & $63.4$ & $24.6$ & $43.6$  \\
     \bottomrule
    \end{tabular}  
    \vspace{-2mm}
    }
    \caption{AP$_{50}$ of models trained with different sizes of sampled patches. Here the number of the sub-patches was set to 16.}
  \label{tab:patchsize}
\end{table}

\subsection{Effectiveness across Object Detectors}
\label{sec:more_detectors}

PBCAT requires no assumption about the structure of the detector and has shown success on the two-stage Faster R-CNN detector in the above experiments. Here we evaluated PBCAT with two additional detectors to validate its effectiveness on more models. Here we used FCOS \cite{fcos}, a typical single-stage object detector, DN-DETR \cite{dndetr}, a transformer-based detector. 


The detectors were trained on MS-COCO. Three methods were evaluated for each detector: standard training, AdvOD, and PBCAT. The patch perturbation step size was $\alpha = 4/255$. Patch perturbation intensity was $\beta = 32/255$ and patch size $\lambda$ was $0.2$. Other settings basically followed those on Faster R-CNN. The detectors with standard training was taken from the \texttt{mmdetection} \cite{mmdetection} repository directly. The evaluation results on FCOS, DN-DETR are shown in \cref{tab:fcos}, \cref{tab:dndetr}, respectively. FCOS trained by PBCAT achieved an AP$_{50}$ of $58.0\%$ on the Inria dataset under the AdvPatch attack, surpassing $l_\infty$-AT by $28.1\%$. Its AP$_{50}$ on AdvTexture and AdvCaT are also much higher than that in Vanilla and $l_\infty$-AT. Similarly, PBCAT significantly improved the robustness of DN-DETR. These results show the effectiveness of PBCAT across diverse object detectors.

\begin{table}[!t]
  \centering
  \small
 \setlength{\tabcolsep}{2.5pt}
  {
      \begin{tabular}{c|c|ccc}
    \toprule
    Method & PGDPatch & AdvPatch & AdvTexture & AdvCaT \\
    \hline
     Vanilla & $17.6$ & $53.9$ & $0.0$ & $0.1$ \\
     
     $l_\infty$-AT (AdvOD) & $26.7$ & $29.9$ & $30.1$ & $17.7$  \\
     
     \rowcolor{lightgray} PBCAT (Ours) & $\mathbf{27.8}$ & $\mathbf{58.0}$ & $\mathbf{55.1}$ & $\mathbf{26.0}$ \\
     \bottomrule
    \end{tabular}   
    \vspace{-2mm}
    }
    \caption{The detection accuracies (AP$_{50}$) of the FCOS models trained with different methods against attacks on different tasks. 
    }
  \label{tab:fcos}
\end{table}

\begin{table}[!t]
  \centering
  \small
 \setlength{\tabcolsep}{2.5pt}
  {
      \begin{tabular}{c|c|ccc}
    \toprule
    Method & PGDPatch & AdvPatch & AdvTexture & AdvCaT \\
    \hline
     Vanilla & $7.3$ & $2.4$ & $0.3$ & $2.8$ \\
     
     $l_\infty$-AT (AdvOD) & $32.0$ & $23.5$ & $0.0$ & $0.1$  \\
     
     \rowcolor{lightgray} PBCAT (Ours) & $\mathbf{32.7}$ & $\mathbf{56.3}$ & $\mathbf{16.8}$ & $\mathbf{56.8}$ \\
     \bottomrule
    \end{tabular}   
    \vspace{-2mm}
    }
    \caption{The detection accuracies (AP$_{50}$) of the DN-DETR models trained with different methods against attacks  on different tasks.}
  \label{tab:dndetr}
\end{table}

\subsection{Robustness in the Physical World}
To further confirm the robustness of PBCAT in real-world scenarios, we evaluated the detector by using four videos provided by the authors of AdvTexture \cite{AdvTexture} and AdvCAT \cite{AdvCaT}. The videos feature two types of texture-based adversarial clothing, tested in both indoor and outdoor environments.  A demonstration video in the Supplementary Materials, compares the detection performance of Faster R-CNN without defense mechanisms and with BCAT. The results indicate that PBCAT significantly improves robustness against physical-world adversarial attacks. Specifically, while the vanilla Faster R-CNN often fails to detect persons wearing adversarial clothing, the model trained with PBCAT successfully detects them in most frames. 

\vspace{-2mm}

\section{Discussion and Conclusion}
\label{sec:cons}

In this work, we introduce PBCAT, a novel adversarial training method to defend against physically realizable attacks. With extensive experiments on the security-critical person detection task, we demonstrated the effectiveness of PBCAT across various scenarios under strong adaptive attacks. Notably, PBCAT demonstrates significant improvements over previous SOTA $l_\infty$-AT method when defending against the AdvTexture \cite{AdvTexture} attack. We encourage future work in enhancing adversarial robustness to consider a broader range of attacks beyond $l_\infty$ attack and patch-based attacks. Additionally, our work highlights that AT is still one of the most promising ways to achieve robustness against physically realizable attacks. The social impact of this work is discussed in \cref{appen:impact}.

\noindent\textbf{Limitation.} 
Similar to most AT works \cite{trades, advod}, PBCAT slightly decreased clean accuracy. It is an open question whether there is an internal trade-off between robustness against physically realizable attacks and clean accuracy. 


\section*{Acknowledgement}
This work was supported by the National Natural Science Foundation of China (No. U2341228).


{
    \small
    \bibliographystyle{ieeenat_fullname}
    \bibliography{main}
}

\appendix

\setcounter{table}{0} 
\setcounter{figure}{0}

\renewcommand{\thefigure}{A\arabic{figure}}
\renewcommand{\thetable}{A\arabic{table}}

\section{Theoretical Cost of Different Patch-Based AT Methods}
\label{appen:computation}
Patch-based AT methods originally designed for classifiers \cite{locationiclr,locationPatchAT,metaAT} incur significant computational overhead when applied to object detectors due to the presence of multiple target objects per image. To quantitatively compare their time complexity, we analyze the number of forward/backward passes required per optimization iteration in \cref{tab:propagation} and provide detailed explanations below. Notably, the theoretical computational overhead is proportional to the number of passes.
\begin{enumerate}
    \item AT-LO \cite{locationPatchAT}: This method selects a single patch per image and performs inference over $k$ steps. In each step, it evaluates $t$ candidate positions and selects the one with the highest loss value. Once the optimal position is determined, a forward/backward pass is conducted to update the patch. When extended to object detection, the computational cost scales linearly with the number of target objects $n$.
    
    \item DOA \cite{locationiclr}: This method also selects a single patch per image. It first performs $t$ steps to determine the patch location, followed by $k$ PGD steps to generate the adversarial patch. The computational cost scales by $n$ when adapted to object detection.

    \item MAT \cite{metaAT}: This method selects $p$ candidate patches, applies them to the image, and computes the loss. It then updates the patch using I-FGSM \cite{i-fgsm}, requiring a total of $p + 1$ steps per iteration. When applied to object detectors, the cost increases proportionally with $n$.

    \item PBCAT (Ours): Unlike the above methods, PBCAT is designed for object detection. It requires only a single forward/backward pass with FreeAT \cite{freeAT}, irrespective of the number of targets, and thus it significantly reduces computational overhead.
\end{enumerate}


\begin{table}[!h]
  \centering
  \small
\setlength{\tabcolsep}{4pt}
  {
      \begin{tabular}{c|c}
   \toprule
    Method & Forward/Backward Times \\
    \hline
     AT-LO \cite{locationPatchAT} & $n \times k \times (1 + t)$\\
     DOA \cite{locationiclr} & $n \times (t + k)$\\
     MAT \cite{metaAT} & $n \times (p + 1)$\\
     PBCAT (Ours) & $1$ \\
     \bottomrule
    \end{tabular}   
    }
  \caption{The number of forward/backward pass required for one optimization step between different patch-based AT methods.}
  \label{tab:propagation}
\end{table}

\section{Pseudo-code of PBCAT}
\label{appen:alg}
The pseudo-code of ``Free'' PBCAT for object detection is shown in \cref{alg:pbcat}.
\begin{algorithm}[!h]
	\caption{
		``Free'' PBCAT on object detection
	}
    \label{alg:pbcat}
	\begin{algorithmic}[1]
		\Require Dataset $\mathcal{D}$, $l_\infty$-bounded global perturbation intensity $\epsilon$, patch perturbation step size $\alpha$, patch perturbation intensity $\beta$, replay parameter $r$, model parameters $\theta$, epoch $N_{\mathrm{ep}}$
		\State Initialize $\theta$
		\State $\mathbf{\delta}_{g} \gets \mathbf{0}$;  $\mathbf{\delta}_{p} \gets \mathbf{0}$; $\delta \gets \mathbf{0}$
		\For{epoch $= 1, \ldots, N_{\mathrm{ep}}/r$}
		\For{minibatch $B \sim \mathcal{D}$}
		\For{i $=1, \ldots, m$}
		\State Compute gradient of loss with respect to $\mathbf{x}$
            \State \qquad $\mathbf{g}_{\mathrm{adv}} \gets \mathbb{E}_{\mathbf{x} \in B}[\triangledown_{\mathbf{x}}\mathcal{L}_d(f_\theta(\mathbf{x} + \delta), y)]$
            \State  Update the model parameter
            \State \qquad $\mathbf{g}_{\theta} \gets \mathbb{E}_{\mathbf{x} \in B}[\triangledown_{\theta}\mathcal{L}_d(f_\theta(\mathbf{x} + \delta), y)]$
            \State \qquad update $\theta$ with $\mathbf{g}_{\theta}$ and the optimizer
		\State Update the patch and global perturbation
            \State \qquad $\delta_{g} \gets \delta_{g} + \epsilon \cdot \mathrm{sign}(\mathbf{g}_{\mathrm{adv}})$
            \State \qquad $\delta_p \gets \delta_p + \alpha \cdot \mathrm{sign}(\mathbf{g}_{\mathrm{adv}})$
            \State \qquad $\delta_{g} \gets \mathrm{clip}(\delta_{g}, -\epsilon, \epsilon)$
            \State \qquad $\delta_p \gets \mathrm{clip}(\delta_p, -\beta, \beta)$
            \State Generate mask $\mathbf{M}$ by the steps in \cref{sec:gradientpatch}
    \State \qquad $\delta \gets \mathrm{Apply}(\delta_p \odot \mathbf{M}, \mathbf{x}) + \delta_{g}$
            
		\EndFor
		\EndFor
		\EndFor
	\end{algorithmic}
\end{algorithm}

\section{Additional Training Settings}
\label{appen:setting}

Following \citet{advod}, we initialized the backbone of the detector using the adversarially pre-trained model provided by \citet{transfer} and trained the detector on MS-COCO for 48 epochs using the AdamW optimizer with an initial learning rate of 0.0001. The learning rate followed a multi-step decay schedule, decreasing by a factor of 0.1 after the 40th epoch. During training, input images were resized such that their shorter side was 800 pixels, while the longer side was at most 1333 pixels.

It is important to note that successful adversarial training on object detectors requires adversarially pre-trained backbones (APB), such as an adversarially trained ResNet-50 on the large-scale ImageNet-1K dataset \cite{imagenet}. However, the backbones of current YOLO-series models are custom-designed networks rather than widely adopted architectures like ResNet-50, and no APB checkpoints are available for these models. Training an APB from scratch on ImageNet-1K is beyond the scope of this work. Nonetheless, we believe that PBCAT could also enhance the robustness of YOLO detectors given sufficient computational resources or available APB checkpoints.

We compared the training time of PBCAT with \citet{advod} in \cref{tab:time}. All experiments were conducted on 8 NVIDIA 3090 GPUs. The slight increase in training time for PBCAT is primarily due to the additional computation required for patch location selection, which is currently implemented on the CPU. This overhead could be further reduced by parallel implementation on the GPU. Overall, we conclude that PBCAT incurs a training cost comparable to standard adversarial training.


\begin{table}[!t]
  \centering
  \small
\setlength{\tabcolsep}{2.5pt}
  {
      \begin{tabular}{c|cccc}
   \toprule
    Method & Faster-RCNN & FCOS & DN-DETR\\
    \hline
     $l_\infty$-AT (AdvOD) \cite{advod} & $34$h & $26$h & $32$h\\
     
     
     PBCAT (Ours) & $44$h & $38$h & $48$h \\
     \bottomrule
    \end{tabular}   
    }
    \caption{The comparison between training time (in hours) of PBCAT and AdvOD.}

  \label{tab:time}
\end{table}

\section{Additional Results and Details}
\label{appen:evaluation}

\subsection{Additional Results on MS-COCO}
\label{appen:coco}

\cref{tab:upstream} shows the results of models trained with various methods on MS-COCO. ``All'' and ``Person'' in \cref{tab:upstream} denote the averaged AP$_{50}$ on each object category and the AP$_{50}$ on the particular person category, respectively.

\begin{table}[!t]
  \centering
  \small
 \setlength{\tabcolsep}{4pt}
  {

    \begin{tabular}{c|cc}
    \toprule
     \multirow{2}*{Method}  & \multicolumn{2}{c}{PGDPatch} \\ 
             & All  & \multicolumn{1}{c}{Person} \\
    \hline
     Vanilla & $18.4$ & $17.5$\\
     
     $l_\infty$-AT (AdvOD) \cite{advod} & $30.7$ & $30.1$\\
     
    \rowcolor{lightgray} PBCAT (Ours) &$\mathbf{37.8}$&$\mathbf{34.5}$\\
     \bottomrule
    \end{tabular}  
    }
    \caption{The detection accuracies (AP$_{50}$) of models trained with various methods on MS-COCO.} 

  \label{tab:upstream}
\end{table}

\begin{figure*}[!t]
\centering
\includegraphics[width=0.8\linewidth]{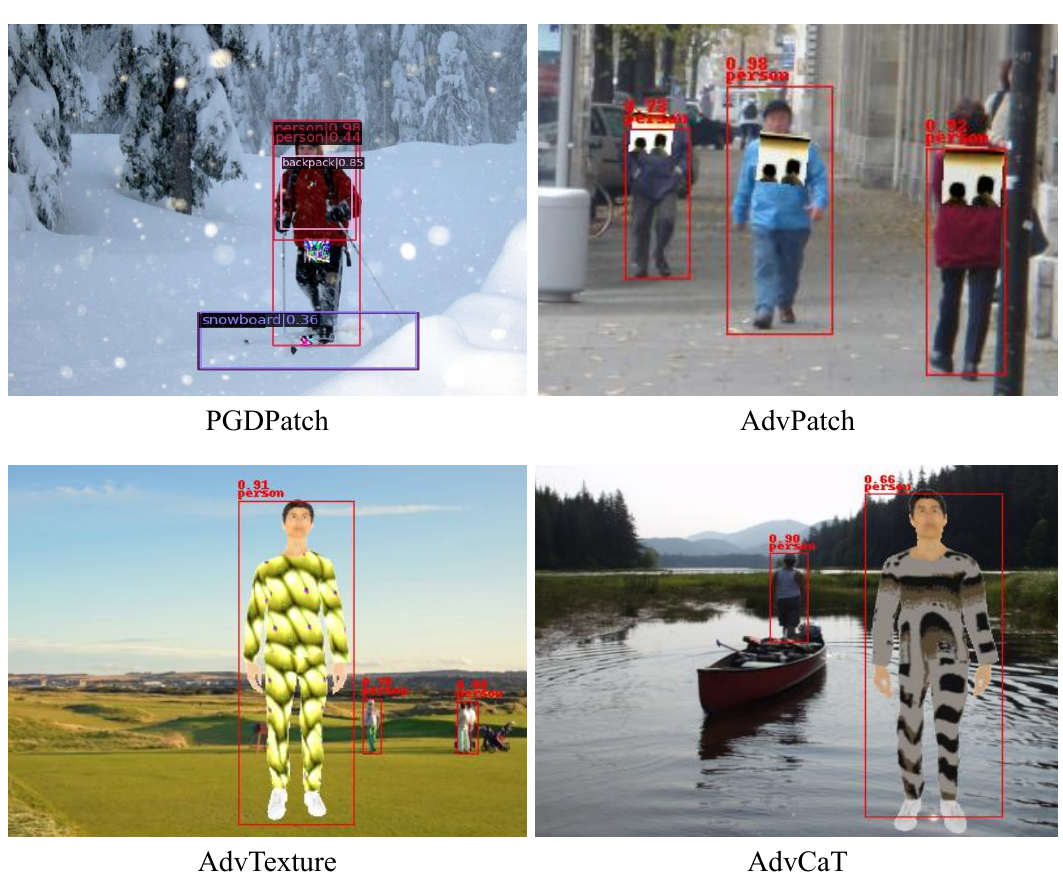}
\caption{The detection results of the model trained with PBCAT under various physically realizable attacks. The detected bounding boxes with confidence larger than 0.5 are visualized.
}
\label{fig:defense}
\end{figure*}

\subsection{Implementation Details of Baselines}
\label{appen:details}
For non-AT defense methods with publicly available source code \cite{epgf, FNC, frame, sac, NAPGuard, jedi}, we downloaded and modified the implementations to conduct adaptive attacks. Specially, we used the official checkpoints provided in the public repositories of SAC\footnote{\url{https://github.com/joellliu/SegmentAndComplete}} and NAPGuard\footnote{\url{{https://github.com/wsynuiag/NAPGaurd}}}
In addition, we used the Matlab code\footnote{\url{https://github.com/ihsenLab/jedi-CVPR2023}} to evaluate Jedi \cite{jedi}.

For non-AT defense methods without released source code \cite{ape, LGS, PatchZero, adyolo}, we reproduced their implementations based on the descriptions in their respective papers.  Note that when reproducing AD-YOLO \cite{adyolo} we replaced its YOLO model with Faster R-CNN for a fair comparison. Other experimental settings closely followed those in the original works.

For SOTA $l_\infty$ AT baseline AdvOD, we directly used the released checkpoint from \citet{advod}\footnote{\url{https://github.com/thu-ml/oddefense}}. For MTD, we reproduced it using the code from \citet{advod} without using adversarially pre-trained backbones.

\subsection{Visualization Results of the Detector}
\label{appen:visualization}
Some visualization results of the Faster R-CNN trained with PBCAT are shown in \cref{fig:defense}. We can see that the detector performed quite well under the large-area and strong adversarial texture attacks.

\begin{table}[!t]
  \centering
  \small

  {
      \begin{tabular}{c|cc}
   \toprule
    Method & T-SEA \cite{T-SEA} & NatPatch \cite{Nat-Patch} \\
    \hline
     Vanilla & $31.7$ & $54.4$ \\
     
     
     PBCAT (Ours) & $\mathbf{90.9}$ & $\mathbf{86.3}$  \\
     \bottomrule
    \end{tabular}  
    }
    \caption{The detection accuracies (AP$_{50}$) of the Faster R-CNN model on transfer-based patch attacks on the Inria dataset.} 

  \label{tab:transfer}
\end{table}

\begin{table}[!t]
  \centering
  \small
  \setlength{\tabcolsep}{2.5pt}
  {
      \begin{tabular}{c|ccc}
      \toprule
    \diagbox{Source}{Target}& Faster-RCNN & FCOS & DN-DETR\\
    \hline
    Faster-RCNN & $77.6$ & $80.7$ & $83.1$\\
    FCOS & $80.0$ & $58.0$ & $79.3$\\
    DN-DETR & $69.2$ & $59.9$ & $56.3$\\
     \bottomrule
    \end{tabular} 
    }
    \caption{The detection accuracies (AP$_{50}$) under AdvPatch on the Inria dataset in the transfer-based attack setting. The adversarial examples generated on the source models were fed into the target models.}  
  \label{tab:blackbox}
\end{table}

\begin{table}[!t]
 \centering
\small
  \setlength{\tabcolsep}{4pt}
 {
   \begin{tabular}{c|ccc}
    \toprule
    Ratio & AdvPatch & AdvTexture & AdvCaT \\  
	\hline
    Top-30\% & $51.3$ & $43.7$ & $52.6$  \\ 
    Top-70\% & $63.0$ & $19.0$ & $43.7$  \\ 
    \rowcolor{lightgray} Top-50\% & $\mathbf{77.6}$ & $\mathbf{60.2}$ & $\mathbf{56.4}$ \\
     \bottomrule
   \end{tabular}
   }
   \caption{The detection accuracies (AP$_{50}$) of models trained with different ratio of sub-patches.}
 \label{tab:selection-topk}
\end{table}

\subsection{Robustness against Transfer Attacks}
\label{appen:transfer}


To further validate that the models trained with PBCAT can have strong robustness against more unseen physically realizable attacks, we additionally evaluated two transfer-based patch attacks \cite{T-SEA, Nat-Patch} on the model trained with PBCAT. The adversarial patches were generated based on their original settings on the vanilla (clean) Faster R-CNN model. The patches were applied to the images in the Inria dataset to evaluate the AP$_{50}$ of Faster R-CNN trained by PBCAT. The results are shown in \cref{tab:transfer}. We can see that these transfer-based patch attacks almost lose the ability to fool the detectors trained with PBCAT.

Additionally,  we used the three types of detectors we trained in this work (as discussed in \cref{sec:more_detectors}), Faster R-CNN, FCOS \cite{fcos}, DN-DETR \cite{dndetr}, to perform the black-box transfer attacks. Here we used the AdvPatch attack on the Inria dataset. The results are shown in \cref{tab:blackbox}. We can also observe that the models trained with our PBCAT can defend these black-box transfer-based attacks better than white-box attacks.

\subsection{Ablation on Top-\textit{k} Selection}
\label{appen:topk}

In \cref{sec:gradientpatch}, we select the top 50\% of sub-patches based on their height values. To investigate the effect of this design choice, we conduct an ablation study varying the top-\textit{k} selection ratios. As shown in \cref{tab:selection-topk}, choosing the top 50\% achieves a strong balance between performance and robustness. Notably, the computational cost remains nearly constant across different selection ratios.

\section{Social Impact}
\label{appen:impact}
Our method increases the robustness of object detectors against physically realizable attacks. This could potentially lead to more effective surveillance systems, which could encroach upon personal privacy if misused. However, we believe the concrete positive impact on security generally outweighs the potential negative impacts. Robust object detectors can enhance various beneficial applications, such as autonomous driving, video surveillance for public safety, and other critical systems. Nonetheless, it is crucial to develop and deploy such technologies responsibly, with ethical considerations to mitigate potential misuse and protect individual privacy.

\end{document}